# Design, Locomotion, and Control of Amphibious Robots: Recent Advances

Yi Jin,[1] Chang Liu[1], Roger D. Quinn,[1] Robert J. Wood,[2*] C. Chase Cao [1,3,4*]

*[1]Department of Mechanical and Aerospace Engineering, Case Western Reserve University, Cleveland, OH 44106, USA*
*[2]John A. Paulson School of Engineering and Applied Sciences, Harvard University, Allston, MA, 02134, USA*
*[3] Department of Electrical, Computer and Systems Engineering, Case Western Reserve University, Cleveland, OH 44106, USA*
*[4]Advanced Platform Technology (APT) Center, Louis Stokes Cleveland VA Medical Center, Cleveland, OH 44106, USA*

*Correspondence authors: rjwood@seas.harvard.edu (RW), ccao@case.edu (CC)*

**Abstract**

Amphibious robots, operating seamlessly across land and water, are advancing applications in conservation, disaster response, and defense. Their performance depends on locomotion mechanisms, actuation technologies, and sensor–control integration. This review highlights recent progress in these areas, examining movement strategies, material-based actuators, and control systems for autonomy and adaptability. Challenges and opportunities are outlined to guide future research toward more efficient, resilient, and multifunctional amphibious robots.

## 1. Introduction

Amphibious robots, designed to traverse both aquatic and terrestrial environments, have emerged as versatile tools in applications ranging from environmental monitoring to emergency response [1-4]. These robots are uniquely positioned to operate in complex and fragile ecosystems, collect critical data with minimal disturbance, and perform search-and-rescue missions in challenging coastal terrains. Natural amphibians, such as turtles, crabs, lobsters, snakes, salamanders, and frogs, have inspired the design of amphibious robots [2]. By emulating their locomotion and adaptive behaviors, researchers have developed a range of bio-inspired driving mechanisms. For example, Baines et al. designed an amphibious robot mimicking the structure and function of turtle flippers [3], while Floyd et al. developed a basilisk lizard-inspired robot capable of walking on land and running on water [4]. Similarly, Yu et al. introduced a snake-like robot driven by body undulation, demonstrating remarkable adaptability across diverse terrains [1].

Beyond bio-inspiration, some researchers have integrated enhancements to improve the performance and versatility of locomotive systems. For instance, Yu et al. designed a fish-inspired amphibious robot equipped with both fins and wheels, where the latter significantly enhanced terrestrial mobility [5]. Other innovative designs include the AmphiHex-II, which features variable stiffness legs for improved adaptability [6], and a rolling robot that utilizes hovercraft-like locomotion for versatile navigation [7]. While existing reviews have discussed various aspects of amphibious robot locomotion, a systematic and detailed classification remains lacking. This paper addresses this gap by offering a comprehensive overview of recent scientific advances, with a focus on locomotion strategies and driving mechanisms.

With recent advances in soft condensed matter physics of active materials, traditional (often rigid) actuators, such as electric motors, have been complemented—and in some cases supplanted—by compliant actuators. These actuators, including dielectric elastomer actuators (DEA) [8], piezoelectric polymers [9], soft magnetic actuators [10], shape memory alloys (SMA) [11], and fluidic elastomer actuators (FEA) [12, 13], have opened new possibilities for amphibious robotics. Compliant actuators offer benefits such as enhanced flexibility, lightweight construction, and biomimetic adaptability, enabling smoother terrestrial/aquatic transitions and a lower Cost of Transport (COT). This review categorizes and evaluates these actuators, examining their mechanisms, performance characteristics, and applicability to amphibious robotics. Additionally, their limitations and prospects are discussed, highlighting avenues for innovation to meet the complex demands of amphibious operations.

Control strategies play a crucial role in the functionality of amphibious robots, particularly in ensuring efficient transitions between terrestrial and aquatic modes. Despite numerous advances in locomotion and control, autonomous mode-switching remains a significant challenge. This review explores advanced sensing and control techniques, emphasizing their role in achieving seamless terrestrial/aquatic transitions and enhancing functional autonomy. As illustrated in **Fig. 1**, locomotion, actuation, sensors, and control strategies form the four foundational pillars of amphibious robot design, which will be discussed in detail in the following sections.

## 2. Locomotion Strategies for Amphibious Robots

Amphibious robots derive their locomotion strategies from natural systems, often taking inspiration from agile amphibians while also extending beyond biological capabilities. These locomotion strategies are specifically designed to enable seamless navigation across both terrestrial and aquatic environments. Locomotion methods can be classified into three primary groups: those optimized for terrestrial movement, those specialized for aquatic environments, and hybrid strategies that function effectively in both environments, as illustrated in **Fig. 2a** and **Fig. 2b**.

### *2.1 Terrestrial Locomotion Strategies*

**Wheeled and Tracked Locomotion.** On land, amphibious robots typically utilize one of three primary locomotion methods: wheeled or tracked systems [14], legged designs [15], and spherical rolling mechanisms[16]. Wheeled and tracked systems rely on frictional forces between the ground and the wheels or tracks to propel the robot forward. These systems are widely regarded as highly energy-efficient, mechanically simple, and capable of carrying substantial payloads, making them the most common choice for land-based locomotion in amphibious robots – as well as terrestrial robots in general. However, their performance is limited on uneven, steep, and rough terrains, where maneuverability decreases significantly. For this reason, engineers build roads and maintain them for efficient rapid-wheeled locomotion.

**Legged Locomotion.** In contrast, legged locomotion provides superior adaptability on complex terrains, where wheeled and tracked systems often falter [17]. Legged robots can generate propulsion through either static friction or controlled slipping. By utilizing alternating foot contact with the ground, legged robots achieve propulsion through static friction between the ground and feet. Furthermore, they can move and climb over irregular surfaces using normal contact between their legs and the substrate, providing much greater ground reaction forces than friction alone can provide. In contrast, controlled slipping relies on deliberately allowing the feet to slide against the surface in a directed manner, rather than maintaining fixed contact [18]. This method can offer enhanced robustness against environmental disturbances [19]. Most importantly, multi-segmented legs allow animals and robots to conform to the rugged terrain [20] and climb over obstacles [21, 22] where wheeled vehicles might overturn. For these reasons, they can navigate obstacles more effectively than wheeled robots. However, the cost is increased by the mechanical complexity of legged systems, requiring more sophisticated control algorithms and actuation mechanisms to maintain balance and efficient movement.

**Spherical Rolling Locomotion.** Another terrestrial locomotion method, spherical rolling, involves the use of a spherical structure to facilitate omnidirectional movement. This method offers advantages such as simplified mechanical construction and the ability to navigate uneven terrain with minimal mechanical wear. However, the spherical shape often limits integration with aquatic locomotion systems, imposing constraints on the overall design and reducing adaptability for hybrid locomotion. A summary of the advantages and disadvantages of these terrestrial locomotion methods is provided in **Table 1.**

### *2.2 Aquatic Locomotion Strategies*

Amphibious robots often employ propulsion methods inspired by marine organisms or engineered propulsion systems in aquatic environments. Common strategies include fin and paddle-based locomotion, water jetting, and propeller-driven propulsion.

**Fin and Paddle-based Locomotion.** Inspired by aquatic creatures such as fish, turtles, and penguins, fin and paddle-based locomotion generates thrust through cyclic appendage movements. Paddle-based locomotion [23] relies on rigid, oscillating structures, while fin-based approaches [24] utilize flexible fins that deform to produce hydrodynamic forces. These methods are mechanically straightforward and energy-efficient, making them well-suited for underwater navigation.

**Water Jetting Propulsion.** Water jetting propulsion [25], inspired by cephalopods and jellyfish, operates by forcefully expelling water through a narrow nozzle to generate thrust, following Newton's third law. This technology enables high burst speeds, excellent station-keeping capability, and rapid directional changes, making it especially effective for evasive maneuvers or agile locomotion.

**Propeller-Based Locomotion.** Propeller-based locomotion commonly used in boats and submarines [26] relies on rotating blades to accelerate water backward, producing continuous thrust. This approach provides high-speed capabilities and improved energy efficiency, making it ideal for sustained underwater propulsion. However, it is less suitable for maneuvering in confined or complex environments than bio-inspired methods. A detailed comparison of the pros and cons of these aquatic locomotion methods is shown in **Table 2**.

### *2.3 Hybrid Locomotion Strategies*

Hybrid locomotion strategies enable amphibious robots to transition seamlessly between land and water by adapting their structural or functional characteristics. These approaches enhance the adaptability and operational versatility of amphibious robots, enabling effective performance across diverse environments. Notable hybrid locomotion methods include body undulation [27], flipper-like legs [28], and rolling hovercraft mechanisms [7].

**Body Undulation.** In terrestrial environments, real-world terrains can be complex, often featuring granular media and obstacles rather than flat ground, conditions referred to as "terradynamics" in prior research [29]. Body undulation is particularly effective in such complex terrains and merits consideration as a viable locomotion strategy [30, 31]. Robots utilizing body undulation mimic the movement of snakes or eels, leveraging flexible, wave-like motions to navigate terrestrial and aquatic environments [32]. This design provides exceptional maneuverability and adaptability, particularly in cluttered or uneven terrains.

**Flipper-Like Legs.** Some amphibious robots integrate leg-like appendages that function as water flippers and legs on land [2]. This design, inspired by amphibians such as turtles and salamanders, provides efficient propulsion across both mediums while maintaining structural simplicity.

**Rolling Hovercraft Mechanisms.** A novel hybrid approach involves rolling hovercraft designs, which combine air-cushion-based propulsion with rolling mechanisms to enable smooth transitions between land and water. This method enhances mobility and adaptability while reducing energy consumption.

**Comparison of Locomotion Strategies.** To evaluate the effectiveness of these locomotion strategies, we conducted a comparative analysis based on key performance metrics, including speed, mobility, and adaptability. The results, illustrated in **Fig. 2c** and **Fig. 2d**, highlight the strengths and limitations of each approach. Wheeled and tracked systems exhibit the highest speeds on smooth terrain due to their efficient energy transfer and continuous ground contact. However, legged locomotion surpasses other methods in terms of mobility and adaptability, enabling robots to traverse uneven surfaces and obstacles more effectively. Propeller-based systems achieve the highest speeds and energy efficiency in water due to continuous thrust generation and streamlined hydrodynamics. In contrast, body undulation provides superior maneuverability and precise directional control, making it advantageous for navigating complex underwater environments. These findings establish a foundation for understanding how locomotion methods are integrated into amphibious robot designs. The following sections will delve deeper into specific applications and innovations associated with these strategies.

## 3. Integrated Designs of Amphibious Robots with Dual Locomotion Strategies

### 3.1 Wheel/track Locomotion-based Amphibious Robots

Wheel and track locomotion is the most prevalent choice for amphibious robots navigating flat, smooth, or structured surfaces due to superior speed, COT, and load-carrying capabilities. This approach forms the foundation for many hybrid designs, where wheels or tracks are paired with aquatic propulsion systems such as fins, paddles, water jets, or propellers to enable seamless operation across terrestrial and aquatic environments.

### ***Wheel/track-fin Locomotion***

Combining wheels or tracks with fins creates a versatile hybrid system capable of operating efficiently on land and in water. For instance, Xia et al. developed a bionic amphibious robot that integrates tracks with side fins to mimic fish-like swimming (**Fig. 3a**) [33]. The robot uses undulating fins for improved buoyancy and maneuverability, achieving speeds of 2 m/s on land and 0.51 m/s underwater. Its parallel mechanism allows smooth transitions between tracks and fins, adapting to various terrains and conditions. Similarly, Yi et al. introduced FroBot, a frog-inspired robot equipped with a universal wheel structure for terrestrial propulsion and flexible caudal fins for aquatic locomotion (**Fig. 3b**) [34]. FroBot achieves underwater speeds of 0.4 m/s, showcasing the potential of integrating fins for dual-environment functionality. Another example is AmphiRobot, developed by Yu et al., which uses wheels for land movement and a caudal fin for propulsion in water (**Fig. 3c**) [5]. This approach demonstrates a simple yet effective design for hybrid locomotion, though challenges remain in balancing performance across mediums, minimizing COT, and refining transition mechanisms.

### ***Wheel/track-paddle Locomotion***

The wheel/track-paddle combination represents an innovative strategy for amphibious robots, enabling efficient operation across diverse environments. In these designs, paddles are often integrated directly into the wheels, enhancing versatility and functionality. For example, an amphibious robot with an Angled Spoke Paddling Wheel (ASPW) incorporates adjustable blades that rotate to generate optimal thrust underwater (**Fig. 3d**) [35]. This mechanism improves propulsion efficiency while maintaining effective land mobility. In another design, the paddle mechanism enhances traction and mobility on soft or uneven terrains by allowing the paddles to dig into the surface (**Fig. 3e**) [23]. This feature also facilitates efficient swimming on water surfaces and underwater. These hybrid systems highlight the adaptability of wheel/track-paddle designs, demonstrating their potential for search-and-rescue missions, environmental monitoring, and other applications requiring seamless transitions between land and water.

***Wheel/track- water-jetting locomotion***

Integrating wheel-based locomotion with water-jet propulsion provides a powerful combination of versatility and performance. Guo et al. developed an amphibious robot incorporating wheels and legs for efficient terrestrial movement while employing independently controlled water-jet thrusters for aquatic propulsion (**Figs. 3f and 3g**) [36]. The robot's design enables it to traverse various terrains in its walking mode and achieve precise maneuverability in water. This hybrid system excels in versatility and adaptability, allowing smooth transitions between environments. However, integrating multiple systems increase mechanical complexity, size, and COT, potentially limiting operational efficiency and duration. Despite these challenges, wheel/track-water jetting designs are highly effective for environmental exploration and disaster response, where multifunctionality is critical.

***Wheel/track-propeller Locomotion***

Wheel/track-propeller designs offer an efficient solution for high-speed operation in terrestrial and aquatic environments. The wheel hub doubles as a propeller in these systems, with mechanisms allowing the wheels to pivot for aquatic propulsion. Boxerbaum et al. developed a robot where the wheels remain parallel to the body for land movement but pivot 90 degrees to activate the propeller hubs underwater (**Fig. 3h**) [37]. Similar designs by Yu et al. use a comparable approach to integrate propulsion directly into the wheels (**Figs. 3i and Fig. 3j**) [38]. This design provides several advantages, including seamless mode transitions, simplified structures, and high-speed operation across environments. However, it also presents challenges, such as increased mechanical complexity, greater energy demands, and higher maintenance requirements due to dual-mode operation. Nonetheless, the wheel/track-propeller combination is a practical and effective solution for amphibious robots requiring speed and efficiency in varied conditions.

### 3.2 Legged Locomotion-based Amphibious Robots

Legged locomotion equips amphibious robots with the ability to navigate challenging terrains such as rocky surfaces, sandy areas, or steep slopes, making these systems invaluable for applications in environmental monitoring, underwater exploration, and search-and-rescue missions. To extend their versatility, legged systems often combine with aquatic propulsion mechanisms, including fins, paddles, and water jets, enabling smooth transitions between land and water.

#### ***Legged-fin Locomotion***

The integration of legs with fins enhances amphibious robots' versatility by combining land mobility with efficient aquatic propulsion. In these designs, legs often serve as the structural framework and driver for fins, coordinating wave-like motions to achieve swimming. For example, as shown in **Fig. 4a** and **Fig. 4b**, robots equipped with side fins utilize leg-driven mechanisms to generate undulating movements for water propulsion [39, 40]. Alternatively, some systems employ independently controlled caudal fins to complement legged locomotion, as illustrated in **Fig. 4c** and **Fig. 4d**. One such robot uses piezoelectric polymers to achieve a jumping motion and reaches a swimming speed of 0.153 m/s [41]. Another design employs a crawling mechanism to move underwater at 0.099 m/s [42]. This hybrid approach leverages the adaptability and mobility of legs for traversing uneven or complex terrains while using fins for controlled and energy-efficient aquatic locomotion. However, combining two distinct locomotion systems introduces challenges, such as increased COT, higher maintenance requirements, and the need for precise coordination between legs and fins. Despite these complexities, legged-fin designs represent an innovative solution for robots operating in diverse and demanding environments.

### *Legged-paddle Locomotion*

Legged-paddle locomotion is a widely used approach that combines the legs' adaptability with the swimming efficiency of paddles. These robots, often called "legged robots with driving feet" [43] utilize distinct paddles that may serve dual purposes: feet for walking on land or as independent propulsion mechanisms for swimming. Based on the number of legs, these robots can be classified into bipedal, quadrupedal, or hexapedal configurations, as shown in **Fig. 5a–d** [4, 44-46]. This design offers several advantages. Legs provide mobility and adaptability on rough terrains, while paddles enhance swimming performance by generating thrust in water. The flexibility to optimize each component for its specific environment ensures efficient performance across varied terrains. For example, bipedal systems are simpler and more agile but less stable, whereas hexapod designs excel in balance and load support but at the cost of increased mechanical complexity and energy requirements. Integrating paddles with legs adds weight and mechanical complexity, and achieving smooth transitions between land and water necessitates precise control systems.

### *Legged-water-jetting Locomotion*

A novel example of legged-water-jetting locomotion is illustrated in **Fig. 5e** and **Fig. 5f**, which depict a spherical amphibious robot equipped with four legs and water-jet thrusters [47]. On land, the robot deploys its legs, mimicking a turtle's gait to walk efficiently. When transitioning to aquatic environments, the robot retracts its legs to minimize drag, while its water-jet thrusters provide propulsion through dedicated openings in the shell. This combination of stable land movement and efficient underwater propulsion demonstrates the versatility of legged-water-jetting designs. The retractable legs and streamlined shell optimize performance in both environments.

However, the spherical design imposes limitations, such as restricted interior space, which reduces payload capacity and leaves less room for additional sensors or components. Despite these constraints, this approach offers a compelling solution for multifunctional amphibious robots.

### 3.3 Spherical Rolling Locomotion-based Amphibious Robots

Spherical rolling locomotion equips amphibious robots with a unique ability to move smoothly across terrestrial surfaces by leveraging their spherical shape. However, the reliance on a rolling mechanism can limit their effectiveness on uneven or steep terrains and constrain the integration of specific aquatic locomotion methods. Common combinations include spherical-rolling-fin and spherical-rolling-propeller locomotion.

#### *Spherical-rolling-fin Locomotion*

An example of spherical-rolling-fin locomotion is shown in **Fig. 6a** and **Fig. 6b**, where an amphibious spherical robot uses a swinging ballast system to induce rolling motion on land [48]. This simple mechanism enables navigation in terrestrial environments with minimal mechanical complexity. The robot maintains its rolling motion in water, with attached fins generating propulsion and steering. This combination simplifies locomotion by eliminating the need for complex leg or wheel systems. However, the reliance on rolling motion reduces the robot's ability to handle uneven terrains effectively, as stability may be compromised. Despite this limitation, spherical-rolling-fin designs are compact, efficient, and well-suited for applications requiring lightweight, dual-environment adaptability, such as environmental monitoring or exploration.

#### *Spherical-rolling- propeller Locomotion*

Spherical-rolling-propeller designs further enhance amphibious mobility by integrating rolling and propeller-driven mechanisms. In the example shown in **Fig. 6c**, the robot achieves terrestrial movement using an internal swinging yoke mechanism that shifts its center of gravity to induce rolling [49]. For aquatic locomotion, integrated propellers generate thrust, enabling efficient swimming. A more advanced design, depicted in **Fig. 6d**, features an inner and outer shell system [50]. The outer shell is driven by wheels positioned between the two layers for controlled rolling on land, while the inner shell houses propellers for aquatic propulsion. These designs demonstrate innovative approaches to achieving seamless transitions between land and water. While spherical-rolling-propeller robots excel in versatility, their mechanical complexity and limited internal space can pose challenges for accommodating additional payloads or components. Nonetheless, such systems offer an elegant solution for amphibious robots requiring robust performance across diverse environments.

### 3.4 Hybrid Locomotion-based Amphibious Robots

Amphibious robots with locomotion systems designed for terrestrial and aquatic environments leverage unique characteristics, structural adaptations, and mode switching to achieve efficient mobility. Such systems include body undulation, flipper-leg, and rolling-hovercraft locomotion, each offering distinct advantages and challenges in navigating diverse terrains and environments.

#### ***Body-undulation locomotion***

Body-undulation locomotion draws inspiration from the wave-like motion of aquatic animals such as sea snakes and eels, enabling amphibious robots to move efficiently in water and adapt this motion for crawling or slithering on land. Robots employing this locomotion style typically utilize electric or softer compliant actuators, such as magnetic actuators, to achieve flexibility and smooth, fluid movement. Examples include the AmphiBot I and II [51, 52] and ACM-R5 [1], illustrated in **Fig. 7a** and **Fig. 7b**, respectively, which rely on snake-like anguilliform undulatory motion for propulsion and crawling. Another example, shown in **Fig. 7c**, is a magnetic-driven amphibious robot capable of crawling on land and undulatory swimming in water, demonstrating its adaptability across diverse terrains [53]. This locomotion style offers excellent maneuverability in complex terrains, such as rocky or muddy surfaces, and efficient, streamlined propulsion in water by minimizing drag. Compliant actuators enhance flexibility while reducing mechanical complexity, making these robots particularly suited for moving through confined or cluttered spaces. However, body-undulation locomotion is not without limitations. Maintaining stability on uneven or slippery surfaces can be challenging, and coordinating multiple segments for smooth undulatory motion increases control demands. Additionally, reliance on compliant actuators may restrict payload capacity and durability in harsh environments.

***Flipper-leg locomotion***

Flipper-leg locomotion combines the functionalities of flippers and legs to enhance versatility and adaptability. Inspired by aquatic animals such as sea turtles and seals, flippers generate powerful thrust for swimming through paddle-like motions. On land, the mechanism transitions to function as legs, enabling walking, crawling, and climbing. Advanced control systems are often integrated to optimize movement efficiency and ensure adaptability to varying conditions. One example is the AmphiHex-I robot, shown in **Fig. 8a**, where the legs transition between the leg and flipper modes by adjusting segment angles [54]. An improved version, AmphiHex-II, illustrated in **Fig. 8b**, incorporates flexible flippers supported by a carbon fiber frame, facilitating seamless transitions between walking and swimming modes [6]. Another example, depicted in **Fig. 8c**, features a turtle-like amphibious robot that transitions between leg and flipper modes using air pressure [3]. This dual-purpose design enhances the robot's versatility, making it suitable for marshes, coastal areas, and underwater ecosystems. However, integrating flipper and leg functionalities may increase mechanical complexity and corresponding control strategies for different types of motions. Additionally, flipper legs may struggle to maintain traction or stability on steep or highly uneven terrains.

***Rolling-hovercraft locomotion***

Rolling hovercraft locomotion represents a unique approach to amphibious mobility, combining efficient terrestrial rolling with smooth hovercraft-like aquatic movement. **Fig. 9a** shows an amphibious robot utilizing three symmetrically arranged ducted fans that provide thrust for rolling motion on land [7]. By adjusting its center of gravity, the robot can transition to hovercraft mode in water, where the bottom surface contacts the water for low-drag movement powered by the fans, as illustrated in **Fig. 9b**. Building on this principle, Yang et al. developed a triphibian robot, shown in **Fig. 9c**, which extends this functionality to aerial locomotion [55]. By rotating the ducted fans to face downward, the robot generates vertical thrust for flight (**Fig. 9d**). Rolling hovercraft locomotion offers significant versatility for land and water operations. The rolling mechanism provides efficient movement on flat or slightly uneven terrains, while the hovercraft mode ensures smooth aquatic mobility with minimal drag. The cylindrical design enhances stability during transitions and simplifies the mechanical structure. However, the system's reliance on ducted fans results in high COT during prolonged operation. Additionally, rolling may be less effective on rough or steep terrains, limiting the robot's applicability.

## 4. Actuators and Actuation Mechanisms for Amphibious Robots

Actuators play a vital role in amphibious robots, working with locomotion systems to enable efficient movement across land and water. This section reviews the actuators commonly used in amphibious robots and their driving mechanisms. Based on structural characteristics, actuators can be broadly classified into rigid actuators**,** such as motors, and compliant actuators**,** including

Dielectric Elastomer Actuators (DEA), Electrohydraulic Actuator (EHA), Piezoelectric Polymer Actuators (PPA), Magnetic Actuators, Fluidic Elastomer Actuators (FEA), Shape Memory Alloys (SMA), Cable-driven and light driven Actuators. While rigid actuators provide reliable power for robust movements like rolling or wheeled locomotion, compliant actuators offer greater versatility and biomimetic capabilities, making them ideal for adaptive and dynamic tasks such as campaniform- or anguilliform-like body undulation or fin-like swimming. These systems collectively enhance the ability of amphibious robots to navigate diverse and complex terrains, as summarized in **Table 3**, which outlines actuator performance, structural characteristics, and design suitability.

### 4.1 Rigid Actuators

Rigid actuators, such as motors, are the most commonly used actuators in amphibious robots due to their high power density, efficiency, and controllability [56]. These actuators effectively drive the robot's end-effector by connecting to one or more joints. For instance, in the AQUA robot, motors control six independently operated flippers, enabling precise movements for both swimming and walking [57]. Similarly, the ArGO Amphibious UGV utilizes motors to operate its wheels, which also function as paddles for water propulsion [58]. Such robust performance ensures that rigid actuators remain essential in the design of amphibious robots. However, challenges persist, especially in micro-robot applications where poor scaling of forces/torques and efficiency can be limiting. Despite these drawbacks, continuous advances in motor technology have improved their efficiency and reliability, solidifying their indispensable role in amphibious robot engineering.

### 4.2 Compliant Actuators

Compliant actuators, made of flexible materials, offer lightweight, low-noise, versatile, and highly adaptable alternatives to rigid actuators. These characteristics have led to their widespread use in amphibious robot design, particularly in applications requiring biomimetic movement and adaptability.

**Dielectric Elastomer Actuators (DEA).** Dielectric elastomer actuators operate using electrostatic forces, with a soft elastomeric material sandwiched between two compliant electrodes. When a voltage is applied, electrostatic forces compress the elastomer in thickness while expanding it in the planar direction, enabling bending and undulatory motion. **Fig. 10a–c** depicts an amphibious robot using DEAs to achieve body bending and locomotion [59]. DEAs offer significant advantages for amphibious robots, including high flexibility, a lightweight structure, and the ability to mimic natural movements like undulation or fin propulsion. These properties make them ideal for navigating narrow spaces and complex terrains and transitioning between land and water. Additionally, DEAs consume relatively low power while producing large deformations. However, challenges such as material durability and limited force output in specific environments remain areas for improvement.

**Electrohydraulic Actuators (EHA).** Electrohydraulic actuators, also known as hydraulically amplified self-healing electrostatic (HASEL) actuators, generate motion by combining electrostatic and hydraulic forces [60]. This actuator is composed of a pouch filled with liquid dielectric and a pair of electrodes printed on the pouch's surface. Applying high voltage to the electrode induces electric field through the liquid dielectric, which leads to Maxwell stress to push the liquid dielectric to working zone. With the volume changing of the dielectric, the hydraulic pressure deforms the actuator structure. **Fig. 10d-f** illustrates a multimodal amphibious robot with three soft electrohydraulic flippers [61]. Compared to traditional dielectric elastomer actuators (DEAs), electrohydraulic actuators (EHAs) can produce larger deformations and achieve faster response times due to the combined action of electrostatic forces and fluid-mediated amplification. However, these advantages come with trade-offs: EHAs are generally more sensitive to environmental factors such as temperature, humidity, and fluid leakage, which can affect their reliability and performance. In addition, they typically require higher operating voltages to initiate and sustain actuation, which may pose challenges for power supply design and long-term durability.

**Piezoelectric Polymer Actuators.** Piezoelectric polymer actuators convert electrical energy into mechanical motion through the converse piezoelectric effect, deforming materials when subjected to an electric field. Known for their precision, rapid response times, and compact size, piezoelectric polymer actuators are well-suited for fine-motion control applications. **Fig. 10g–i** illustrates a micro amphibious robot using piezoelectric polymers for leg and caudal fin motion, where voltage differences control bending and propulsion [41]. While these actuators excel in generating efficient oscillatory propulsion in aquatic environments, their limited range of motion necessitates amplification mechanisms for larger-scale movements.

**Magnetic Actuators.** Magnetic actuators leverage electromagnetic fields, magnetic particles, and magnetic cores to drive locomotion. **Fig. 10j and k** shows an amphibious robot actuated by a magnetic coil system, which enables walking and swimming through magnetic interactions [53]. Compared to other compliant actuators, magnetic actuators reduce the robot's size and provide a fast response, making them advantageous for confined environments. However, their reliance on external magnetic fields limits operational flexibility, and further advancements are needed to expand their applications.

**Fluidic Elastomer Actuators (FEA).** Fluidic elastomer actuators are driven by air or fluid pressure and consist of channels in an elastomeric matrix, gas or hydraulic valves, and associated plumbing [62]. These actuators are frequently used in amphibious robots for transitions between modes or to control the state of deformable limbs. For example, **Fig. 11a** shows an FEA used for mode transitions [13], while **Fig. 11b–c** depicts deformation-driven actuation [12]. While FEAs offer fast response times and adaptable motion, they need additional equipment to generate and control the fluid distribution with suitable pressures, which increases COT and makes them challenge to implement untethered robots. Their energy efficiency is also a notable limitation, and energy transformation rates typically peak at around 50%, with losses stemming from heat dissipation, fluid friction, and elastic deformation. Additionally, buoyancy effects caused by trapped air or fluid can significantly influence performance, either aiding or resisting intended motions, depending on the system design.

**Shape Memory Alloys (SMA).** Shape memory alloys are actuators capable of transitioning between predefined shapes when exposed to different temperatures. **Figure 11d** shows an SMA integrated into an amphibious robot's leg, with its deformation stages (austenite, twinned martensite, and detwinned martensite) illustrated in **Fig. 11e** [63]. SMA-driven robots can swim or crawl by repeatedly heating and cooling the material, using conduction-based water-cooling processes. However, cooling on land remains challenging, slowing operational speed and limiting SMA applicability in terrestrial environments.

**Cable-driven Actuators.** Cable-driven actuators are commonly used in amphibious robots, particularly those employing undulatory locomotion. By adjusting the length of the cables, these robots can bend or contract their bodies, generating wave-like movements for propulsion. Fig. 11f and g illustrate a snake-like amphibious robot, named AquaMILR+, which utilized variable-stiffness cables to navigate effectively through complex aquatic environments [64]. However, cable-driven amphibious robots face drawbacks, including friction and backlash within the cable mechanisms.

**Light-driven Actuators.** Light-driven Actuators. Light-driven actuators convert optical energy into mechanical motion through mechanisms such as photothermal expansion, photochemical reactions, or photomechanical deformation in materials including liquid crystal elastomers (LCEs), carbon-based composites, and photothermal hydrogels. They offer distinct advantages such as wireless and remote activation, high spatial precision, and the ability to trigger motion without direct physical connections, making them particularly attractive for lightweight and untethered systems. In amphibious robots, light-driven actuators can enable versatile tasks—such as underwater crawling, swimming, or terrestrial locomotion—by selectively illuminating different regions to induce bending, twisting, or undulatory motions. This non-contact control is especially valuable in confined or hazardous environments. However, significant challenges remain, including achieving sufficient force output, maintaining fast response in aqueous conditions, and ensuring reliable operation under variable light intensities. Figures 11h and 11i illustrate a light-driven amphibious robot. Despite their promise, these actuators are limited to transparent environments, and photothermal mechanisms are generally inefficient in water due to its high thermal capacity, which dissipates absorbed heat [65].

## 5. Control Strategies for Amphibious Robots

Amphibious robots, designed to operate seamlessly on land and in aquatic environments, face unique challenges in control strategy development. Effective control strategies integrate robust sensor fusion, real-time decision-making algorithms, and dynamic models to manage transitions between terrestrial and aquatic modes while maintaining stability and efficiency. However, performance can be significantly affected by sensor noise and environment-induced disturbances. On land, vibration, terrain irregularities, or wheel-ground slip may introduce errors in proprioceptive and exteroceptive sensing, complicating state estimation and locomotion control. In aquatic environments, factors such as water turbidity, light refraction, currents, and surface waves can distort sensor readings, reduce visibility, and cause unmodeled external forces. These uncertainties demand adaptive filtering techniques and disturbance rejection mechanisms to ensure reliable perception, robust control, and consistent performance across different conditions.

### 5.1 Control strategies

Amphibious robots navigate dynamic and diverse environments, necessitating adaptive and robust control methodologies to ensure smooth transitions between modes, stable operation, and efficient task execution. Over time, various control approaches have been developed, each tailored to address specific challenges.

**PID Control.** Proportional-integral-derivative (PID) control is widely used to maintain stability by minimizing errors between desired and actual states. This method adjusts control inputs based on proportional, integral, and derivative terms to achieve precise control [66, 67]. For instance, the AmphiBot-I robot employs PID control to ensure stable body undulation during swimming and crawling [51]. While effective in predictable environments, PID control struggles in highly dynamic or nonlinear conditions, limiting its versatility. For instance, an amphibious robot using PID control could struggle when transitioning from water to land due to the drastic change in dynamics. Similarly, the PID controller may not respond quickly in highly dynamic aquatic environments with varying currents and turbulence. In addition, sensor noise (e.g., IMU drift, camera distortion, or pressure sensor fluctuations) and environment-induced disturbances can introduce false error signals that cause oscillations, overshooting, or sluggish responses in the controller. To mitigate these issues, PID tuning must be carefully adapted—using low-pass filtering or sensor fusion to reduce noise effects, derivative filtering to prevent amplification of high-frequency disturbances, and adaptive or gain-scheduled tuning to account for different operating regimes between land and water. In these scenarios, PID control may fail to provide precise adjustments, leading to instability and poor performance.

**Model Predictive Control (MPC).** MPC optimizes control inputs over a finite prediction horizon using a dynamic system model [68-70]. It has proven effective in tasks requiring precise movement, such as optimizing fin and limb coordination in amphibious robots. However, its high computational demands often restrict real-time applications, particularly in resource-constrained systems. Furthermore, MPC performance can degrade when sensor noise or environment-induced disturbances corrupt state estimations or prediction models. For instance, inaccurate position or velocity data from noisy sensors may propagate through the prediction horizon, leading to suboptimal control actions or constraint violations. To address these issues, disturbance observers, robust or stochastic MPC formulations, and sensor fusion with noise filtering can be employed to enhance resilience.

**Finite State Machines (FSM).** FSMs provide a structured framework for managing operational modes and state transitions through predefined rules [71]. Each state corresponds to a specific mode, such as walking, swimming, or switching between the two. FSMs excel in predictable environments where transitions can be discretely defined. However, their reliance on predefined logic makes them less adaptable in unstructured or unpredictable conditions.

**Reinforcement Learning (RL).** RL enables robots to learn optimal control policies through trial-and-error interactions with the environment. This approach is particularly advantageous in uncertain or complex environments that are difficult to model. For example, the Amphibious

Wheel-Legged Robot (AWLR) employs RL to optimize locomotion strategies, facilitating effective transitions between wheeled and legged propulsion [72]. However, RL requires extensive training, computational resources, and a well-designed reward structure, which limits its practicality for real-time deployment.

**Behavior-Based Control.** Behavior-based control decomposes complex tasks into modular behaviors, such as obstacle avoidance, path-following, or speed regulation [47]. These behaviors function independently but are orchestrated to achieve overarching goals. While effective in modular task execution, this approach can suffer from behavior conflicts when multiple behaviors are triggered simultaneously, leading to inefficient or unpredictable responses. Additional strategies, including bio-inspired control [73] and adaptive control [74], have also been applied to amphibious robots. **Table 4** summarizes the performance and applications of various control strategies in amphibious robotics.

### 5.2 Sensing Feedback

Sensing systems are integral to amphibious robots, enabling environmental perception, balance, position monitoring, and force/torque measurement across terrestrial and aquatic domains. Sensors can be broadly categorized into three groups: those optimized for land, those specialized for underwater use, and those functioning effectively in both environments.

**Land-Based Sensors**: These include antennas [75], infrared sensors [76], and accelerometers [77], which have traditionally been developed for operation in terrestrial environments.

**Underwater Sensors**: These are tailored for aquatic use, such as water detectors, flow sensors [78], sonar [79], and pressure sensors [80], enabling effective navigation and monitoring in submerged environments.

**General Sensors for Both Environments**: These include cameras [81, 82], compasses [83], GPS [84], and angle sensors [80], offering versatility for amphibious robots to perform tasks in both domains.

The combination of these sensors equips amphibious robots with the ability to perceive their surroundings, maintain balance, and monitor forces and positions, ensuring efficient operation in complex and dynamic environments.

### 5.3 Integrating Control and Sensing for Mode Transition

A critical aspect of control strategies in amphibious robots is their integration with sensing systems to facilitate mode transitions between terrestrial and aquatic environments. These transitions involve significant shifts in environmental dynamics, requiring precise coordination between sensory input, control algorithms, and actuator responses. Effective integration ensures smooth mode switching and enhances adaptability to varying terrains and water conditions, ultimately determining the robot's practicality and reliability in real-world applications. **Table 5** lists examples of amphibious robots and their control strategies for mode switching.

### 5.4 Hydrodynamics and Robot–Fluid Coupling in Control Design

Effective control of amphibious robots requires not only robust algorithms but also an understanding of the hydrodynamic environment in which they operate. Aquatic locomotion is inherently influenced by fluid–structure interactions, added mass effects, vortex shedding, and unsteady drag, all of which complicate control tuning and validation [85]. Traditional control strategies such as PID often rely on empirical tuning to stabilize swimming or crawling gaits, but their performance can degrade under fluctuating flow conditions and disturbances [86]. Advanced approaches, including MPC and RL, have been investigated to improve adaptability, yet their effectiveness depends strongly on accurate representations of robot–fluid dynamics [87, 88].

Computational fluid dynamics (CFD) simulations and reduced-order hydrodynamic models provide a critical foundation for evaluating these interactions. Navier–Stokes–based solvers and immersed boundary methods have been used to study propulsion mechanisms such as fin undulation, jet expulsion, or limb paddling, offering insights into thrust generation, efficiency, and wake structures [89, 90]. Such modeling frameworks also allow virtual testing of control strategies before physical implementation, reducing experimental cost and risk. Coupling CFD results with control algorithms (e.g., embedding drag or added-mass terms into MPC cost functions) has been shown to improve trajectory tracking and disturbance rejection [69, 91].

In amphibious contexts, hybrid locomotion presents additional complexity: transitions between terrestrial and aquatic domains involve rapid changes in contact forces, buoyancy, and drag. Thus, fluid–structure coupling models are particularly valuable for designing controllers that can adapt across modes [92]. Incorporating these models into reinforcement learning frameworks further enables policy training in simulated environments that account for fluid dynamics, enhancing transferability to real-world conditions [93].

Despite these advances, challenges remain in balancing fidelity and computational cost. High-resolution CFD simulations are resource-intensive, limiting their direct integration with real-time control. Emerging efforts focus on surrogate modeling, reduced-order representations, and physics-informed machine learning to bridge this gap [94, 95]. These approaches point toward a future where hydrodynamic modeling and control design are more tightly integrated, enabling amphibious robots to operate robustly across diverse aquatic environments.

## 6. Challenges & Outlook of Amphibious Robots

Despite significant advances in amphibious robotics, several challenges persist, limiting the full potential of these systems. These challenges span design, actuation, control strategies, and environmental interaction. However, they also present opportunities for innovation and technological advancement, enabling the development of more efficient, adaptable, and robust amphibious robots.

### 6.1 Challenges

**Design Challenges**: The design of amphibious robots may be improved by integrating terrestrial and aquatic locomotion mechanisms. While combining these systems enables dual-environment operation, it also introduces several complexities. Many existing amphibious robots rely on separate land and water movement mechanisms, each requiring independent control strategies and power sources, which increases COT and structural complexity. Although hybrid systems, such as the Angled Spoke Paddling Wheel (ASPW), simplify control by embedding paddles into wheels and efficiently combining terrestrial and aquatic systems, locomotion remains underdeveloped. Furthermore, many amphibious robots draw inspiration from biological organisms, mimicking the undulating propulsion of fish or the salamander-like gait of amphibians. While bioinspired designs demonstrate effective locomotion, they often fail to perform optimally in real-world environments with soft sand, slippery algae, or sharp rocks. A significant challenge lies in developing practical designs that balance bio-inspired efficiency with engineered robustness.

**Actuation Challenges**: Actuators are critical in amphibious robots, as they directly influence locomotion, adaptability, and environmental interaction. Rigid actuators (e.g., motors) are widely used due to their ubiquity, ease of use, high power density, and efficiency. However, they suffer from heavy weight, rigidity, and poor scalability, which limits the adaptability of the amphibious systems in dynamic environments. Compliant actuators, which utilize stimuli-responsive materials such as electroactive polymers or shape-memory alloys, offer advantages in flexibility and lightweight construction. However, they face limitations in load-carrying capacity, control precision, and response time. Hybrid actuators, which integrate rigid and soft actuation mechanisms, are needed to bridge the gap between strength and adaptability, enabling seamless movement across diverse terrains.

**Control Challenges:** Controlling amphibious robots is complex due to the need for smooth transitions between terrestrial and aquatic environments. Seamless mode-transition requires advanced control algorithms that adapt to real-time environmental changes, such as terrain variations (e.g., soft sand, muddy surfaces, turbulent water current), complex fluid dynamics (e.g., drag forces and resistance changes), and structural stresses during mode-switching. Technologies like Finite State Machines (FSM) and Reinforcement Learning (RL) offer promising frameworks to provide robust control adaptability in the presence of unknown disturbances. However, these methods often struggle when dealing with multiple variables simultaneously, making transition less efficient. Additionally, sensor limitations present another major hurdle. Underwater environments introduce challenges such as low visibility, signal interference, and cluttered surroundings. Traditional rigid sensors that cannot deform may fail to provide conformal interaction and real-time accurate feedback during mode transitions, reducing the robots' ability to adapt dynamically to their surroundings.

## 6.2 Future Prospects

While these challenges pose significant obstacles to developing amphibious robots, they also open exciting opportunities for technological advances in robot design, actuation, and control systems.

**Innovative Design Approaches:** To address the complexities of integrating terrestrial and aquatic locomotion, researchers are developing hybrid designs that combine multiple propulsion mechanisms. For example, combining wheels, paddles, or propellers can enable smooth transitions between land and water. Artificial intelligence (AI)-driven generative design and machine learning algorithms can simulate robot-environment interactions to generate optimized, adaptive structures [96, 97]. Incorporating compliant mechanisms and soft materials into robot designs enables conformal adaptation to uneven or dynamic surfaces (e.g., rocky shores or muddy terrain) for enhanced stability and energy efficiency. For example, Grezmak et al. developed a sensory leg with a compliant end-effector for an amphibious crab robot [98]. Using embedded magnets and magnetometers, the leg could sense terrain compliance and adjust locomotion accordingly. Such terrain-aware designs improve navigation efficiency in unpredictable environments. Additionally, integrating aerial capabilities into amphibious robots could enhance their versatility in complex environments. Current research explores flipper-wing hybrids that allow micro-robots to transition between water and air [99, 100]. These designs merge aquatic locomotion (flippers) with aerial lift mechanisms (wings), enabling multimodal functionality.

**Advanced Actuation Systems:** Developing hybrid actuators that combine rigid precision with soft adaptability is key to overcoming the limitations of existing systems. Future actuation innovations may include:

- Compliant actuators with tunable stiffness: Materials that adjust their stiffness based on environmental conditions to improve energy efficiency and force output.
- Bio-inspired muscle actuators: Soft robotic actuators that mimic natural muscle contractions for adaptive movement in dynamic terrains.

- High-efficiency electrostatic actuators: Offering lightweight, high-speed actuation while maintaining low power consumption for prolonged operational endurance.
- Biohybrid actuators: Combine living biological components (such as muscle cells and tissues) with synthetic materials (like hydrogels or polymers) to develop actuators that can move, bend, or contract in response to stimuli [101-103].

These advancements will enhance locomotion performance, allowing robots to navigate seamlessly across various terrains while maintaining high energy efficiency.

**Intelligent Control Systems:** Integrating AI-enhanced control strategies with advanced sensing technologies [104-108] will improve the adaptability of amphibious robots. Potential future developments include:

- AI-enhanced sensor fusion algorithms: Combining data from IMUs, cameras, and environmental sensors to create real-time environmental maps.
- Soft sensors for enhanced terrain feedback: Using flexible materials (e.g., hydrogels) to measure force, strain, and pressure, enabling precise locomotion adjustments.
- Reinforcement Learning (RL) for dynamic navigation: Allowing robots to continuously adapt their locomotion based on real-world environmental feedback.

Future control architectures will leverage adaptive algorithms to optimize mode transitions, reduce energy consumption, and enhance locomotion efficiency in complex hybrid terrains. Such advances will enable amphibious robots to perform effectively in real-world applications, such as environmental monitoring, disaster response, and scientific exploration.

## 7. Conclusion

Recent advances in soft materials, advanced control strategies, and cutting-edge sensing and fabrication technologies have significantly enhanced the capabilities of amphibious robots, enabling exceptional performance and expanding their application potential across various fields. This paper provides a comprehensive review of the state-of-the-art of amphibious robotics, focusing on their locomotion mechanisms in aquatic and terrestrial environments, actuator design and actuation strategies, control methods, and associated challenges. Despite these reported advances, several limitations persist, particularly in 1) integrating terrestrial and aquatic systems while maintaining efficiency and adaptability, 2) enhancing actuator performance to balance strength, flexibility, and energy efficiency, and 3) refining control strategies for seamless mode-transition across different environments. Addressing these challenges will require innovative solutions such as 1) AI-assisted design frameworks to systematically optimize amphibious robot morphology and functionality; 2) developing novel soft sensors that enhance environmental perception and adaptive control to enable robots to navigate complex terrains and transition zones more effectively; 3) innovating new soft materials and compliant mechanisms that improve adaptability, durability, and energy efficiency, ensuring optimal performance in dynamic and unpredictable environments. Looking ahead, the continued integration of advanced materials, actuation technologies, and AI-driven control systems will unlock the full potential of amphibious robots. These advances will pave the way for transformative applications in environmental monitoring, search-and-rescue operations, disaster response, underwater exploration, and beyond. By overcoming limitations and leveraging future innovations, amphibious robots will become more autonomous, efficient, and versatile, expanding their role in real-world missions across diverse and challenging terrains.

**Acknowledgments**

This work was partially supported by the National Science Foundation (ECCS-2024649), USDA NIFA (Grant No. 2021-67021-42113), and Case Western Reserve University. The authors thank Dr. Kathryn Daltorio and Jordan Gray at Case Western Reserve University for their valuable discussions.

**Author contributions**

Y.J. conducted the literature review and drafted the initial manuscript. C.L. contributed to the editing of the manuscript. R.Q. and R.J.W. provided critical insights into integrating bioinspired design and control strategies and edited the manuscript. C.C.C. conceived and the project and contributed to drafting and editing the manuscript. All authors discussed and approved the manuscript.

**Competing Interests**

The authors declare no competing interests.

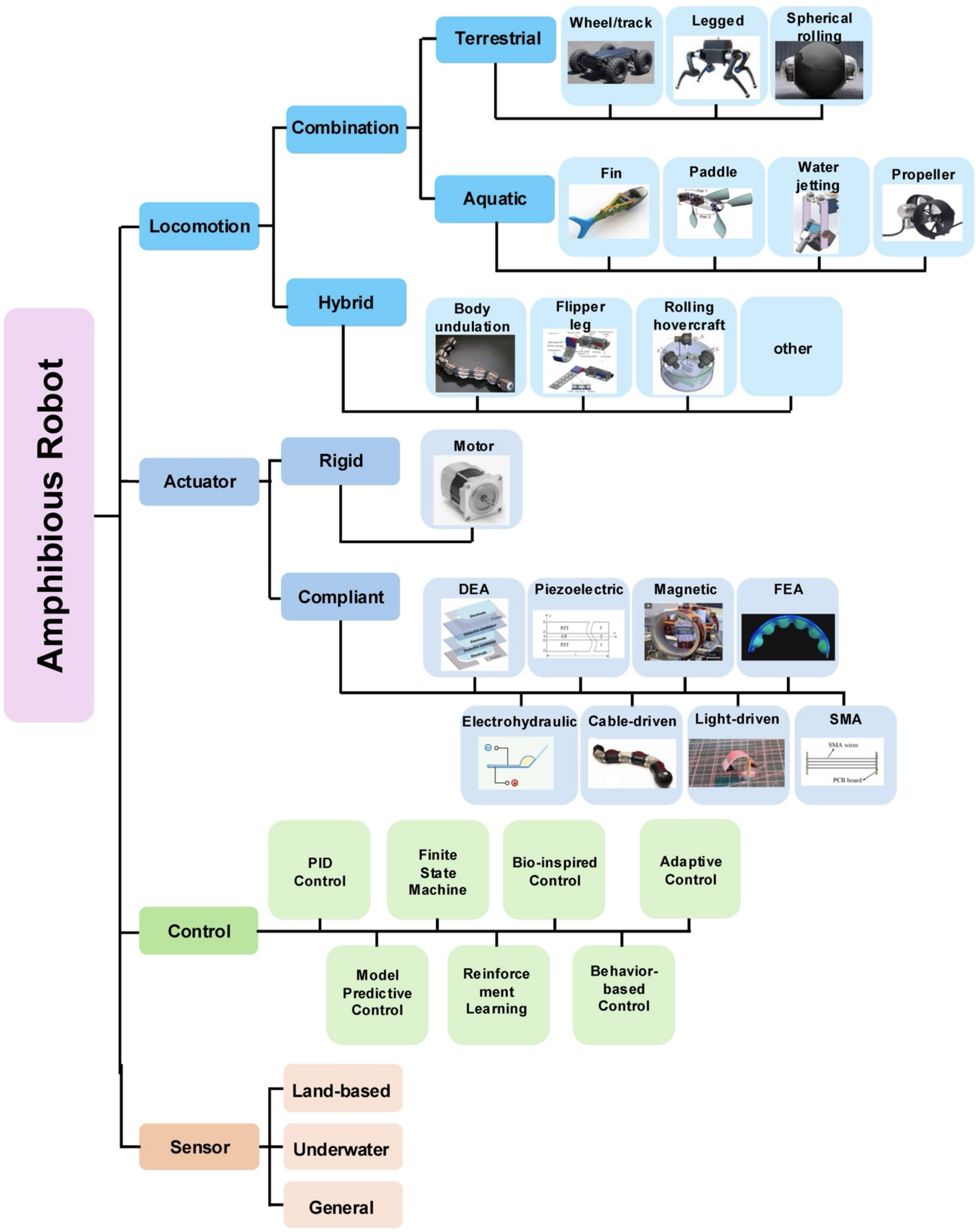

**Fig. 1. Design factors of amphibious robots.** The three key factors influencing amphibious robot design are locomotion mechanisms, actuation systems, and control strategies. Locomotion

integrates terrestrial and aquatic methods for seamless transitions; actuation provides power and motion through rigid or soft systems; and control strategies ensure stability and adaptability in complex environments.

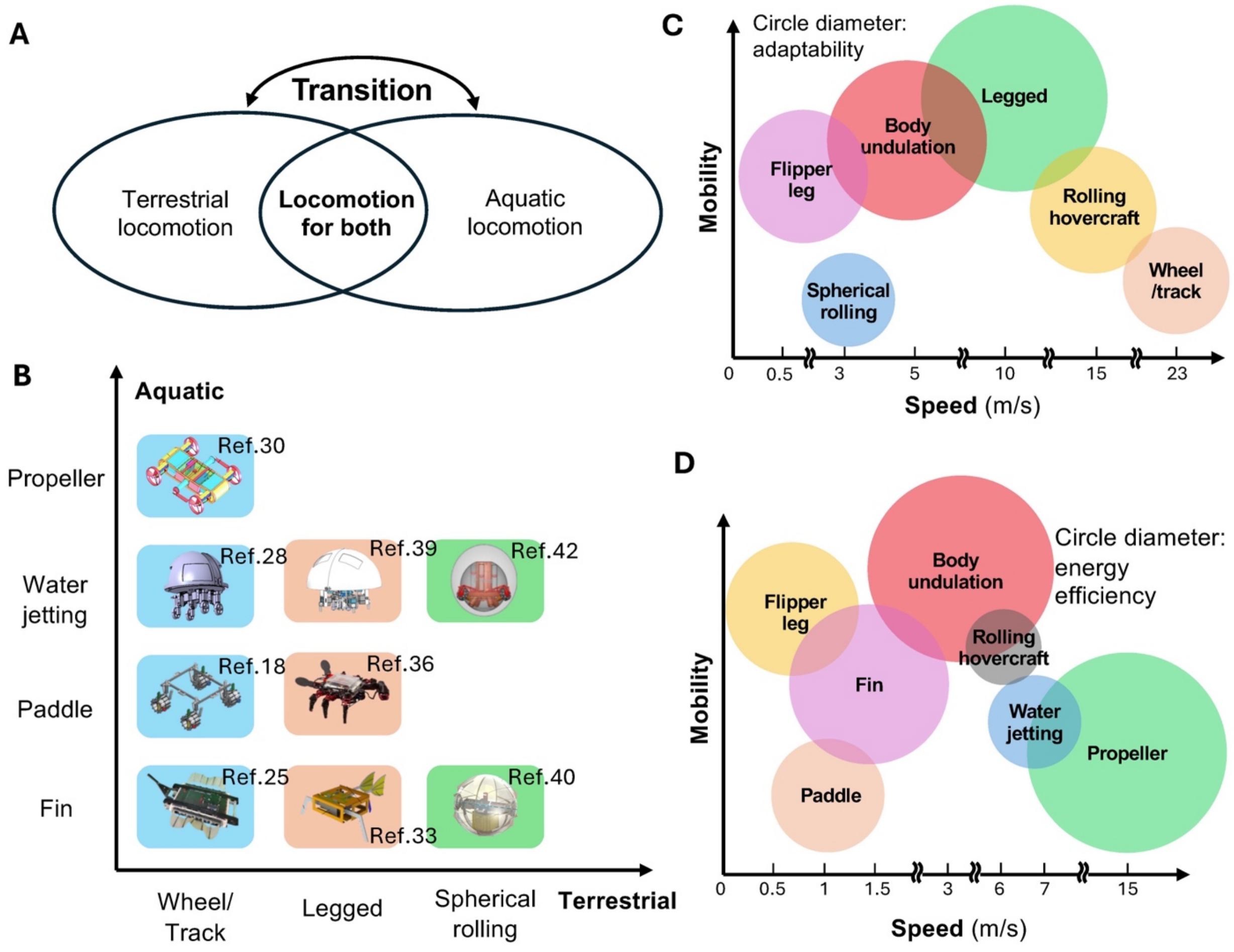

**Fig. 2. Locomotion strategies and performance comparisons of amphibious robots. (A)** Two locomotion strategies: combining terrestrial and aquatic locomotion systems or using unified locomotion mechanisms for both environments. (**B)** Examples of different combinations of terrestrial (e.g., wheels, legs) and aquatic (e.g., fins, paddles) locomotion systems. (**C)** Comparison of terrestrial locomotion methods based on speed, mobility, and adaptability. (**D)** Comparison of aquatic locomotion methods based on speed, mobility, and energy efficiency.

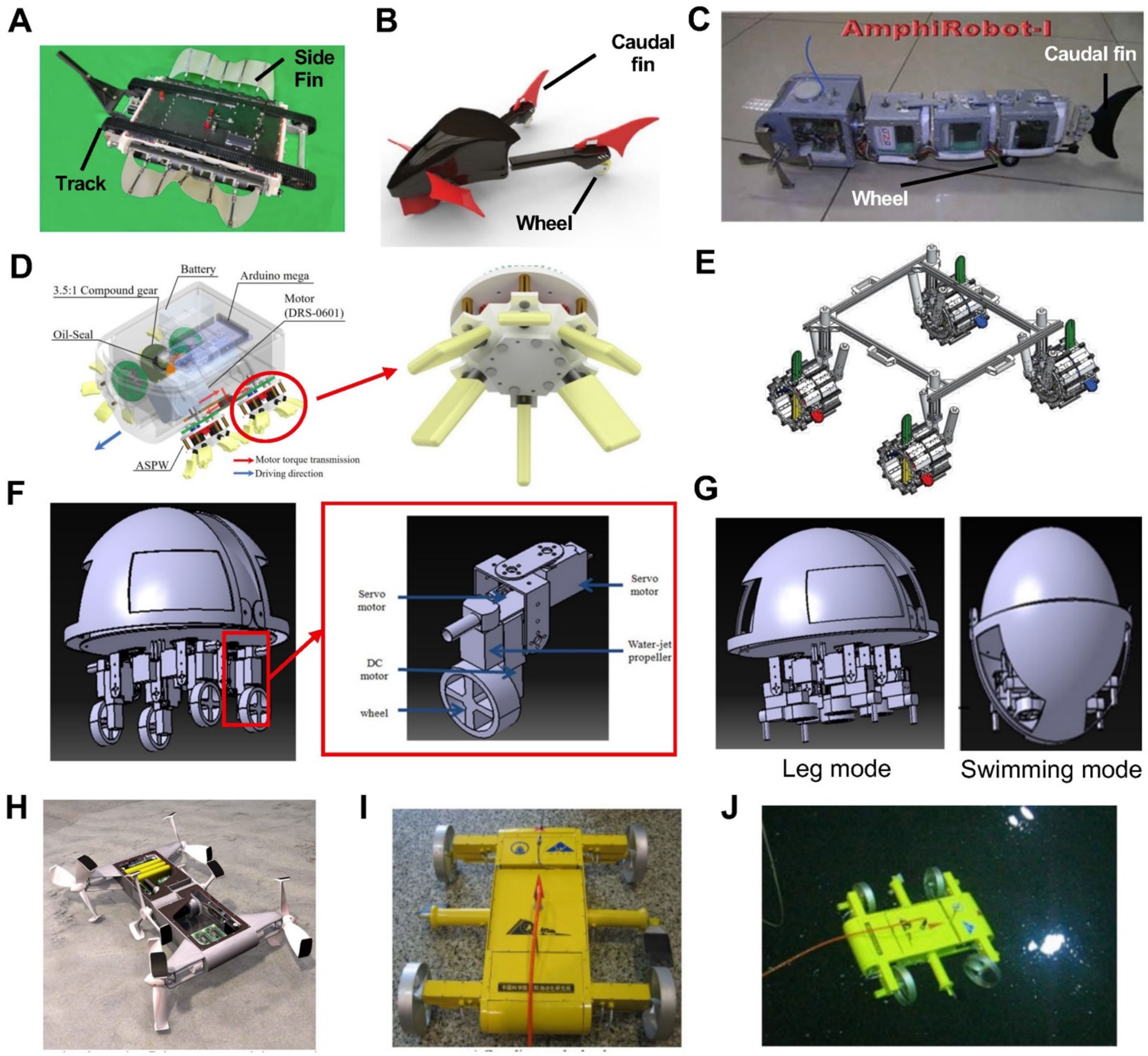

**Fig. 3. Wheel/track-driven amphibious robots.** (A–C) Amphibious robots utilizing wheel/track-fin locomotion for dual-environment adaptability [5, 33, 34]. (D–E) Robots employing wheel/track-paddle locomotion for enhanced terrestrial and aquatic efficiency [24, 32]. (F) Schematic of a wheel/track-water-jetting robot, enabling powerful aquatic propulsion [36]. (G) Transition to leg mode for navigating rough terrain and swimming mode for aquatic movement. (H–J) Schematic of wheel-propeller integrated robots for seamless operation across land and water [37, 38].

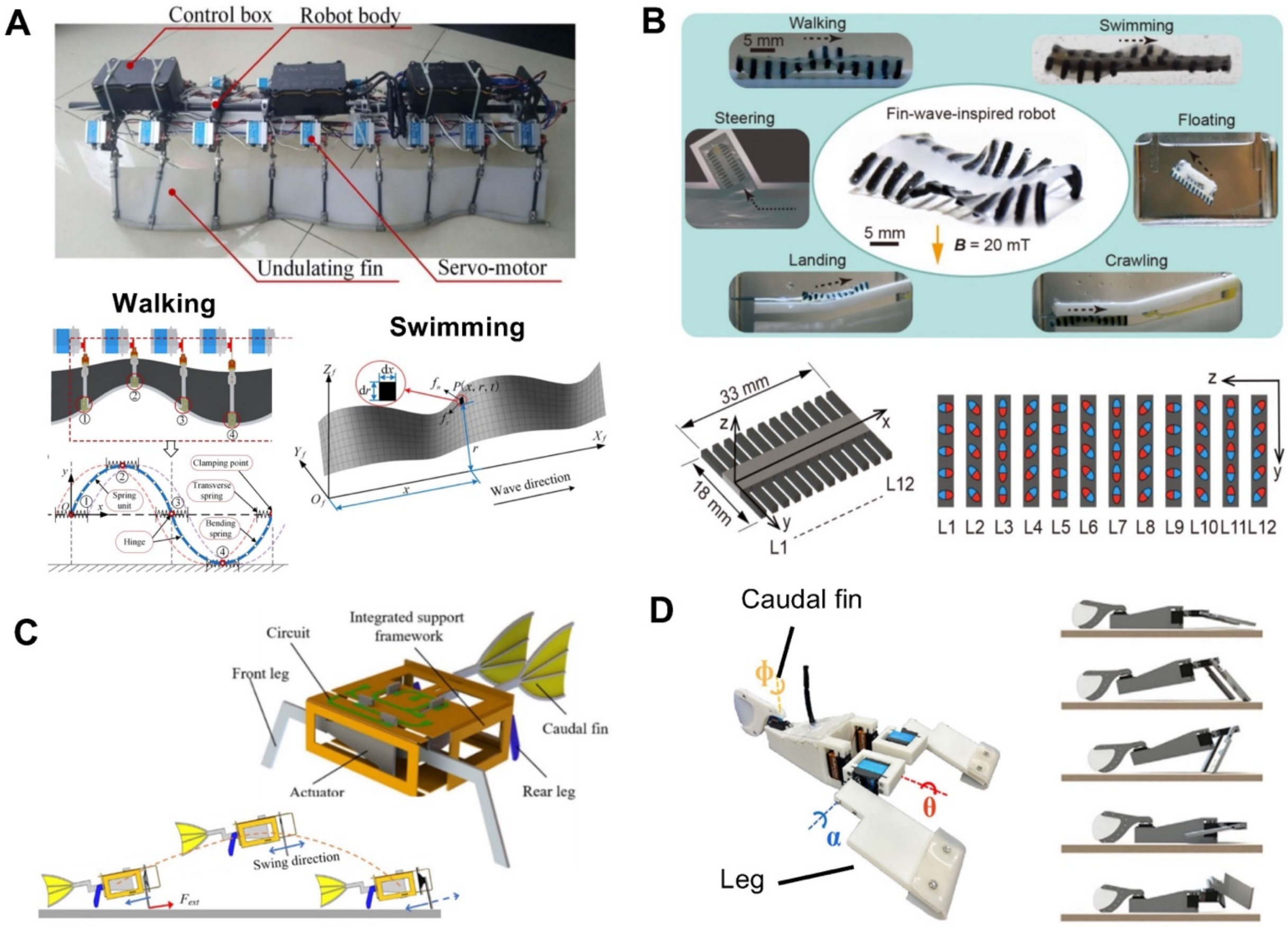

**Fig. 4. Legged-fin locomotion in amphibious robots. (A)** Amphibious robot with a composite wave fin for efficient aquatic propulsion [39]. **(B)** Fin-wave-inspired magnetic soft robot demonstrating flexible undulatory motion [40]. **(C)** SJTU Amphibious Microrobot driven by piezoelectric polymers for precise leg and fin motion [41]. (**D)** Schematic illustration of MIARF, integrating leg and fin for versatile amphibious locomotion [42].

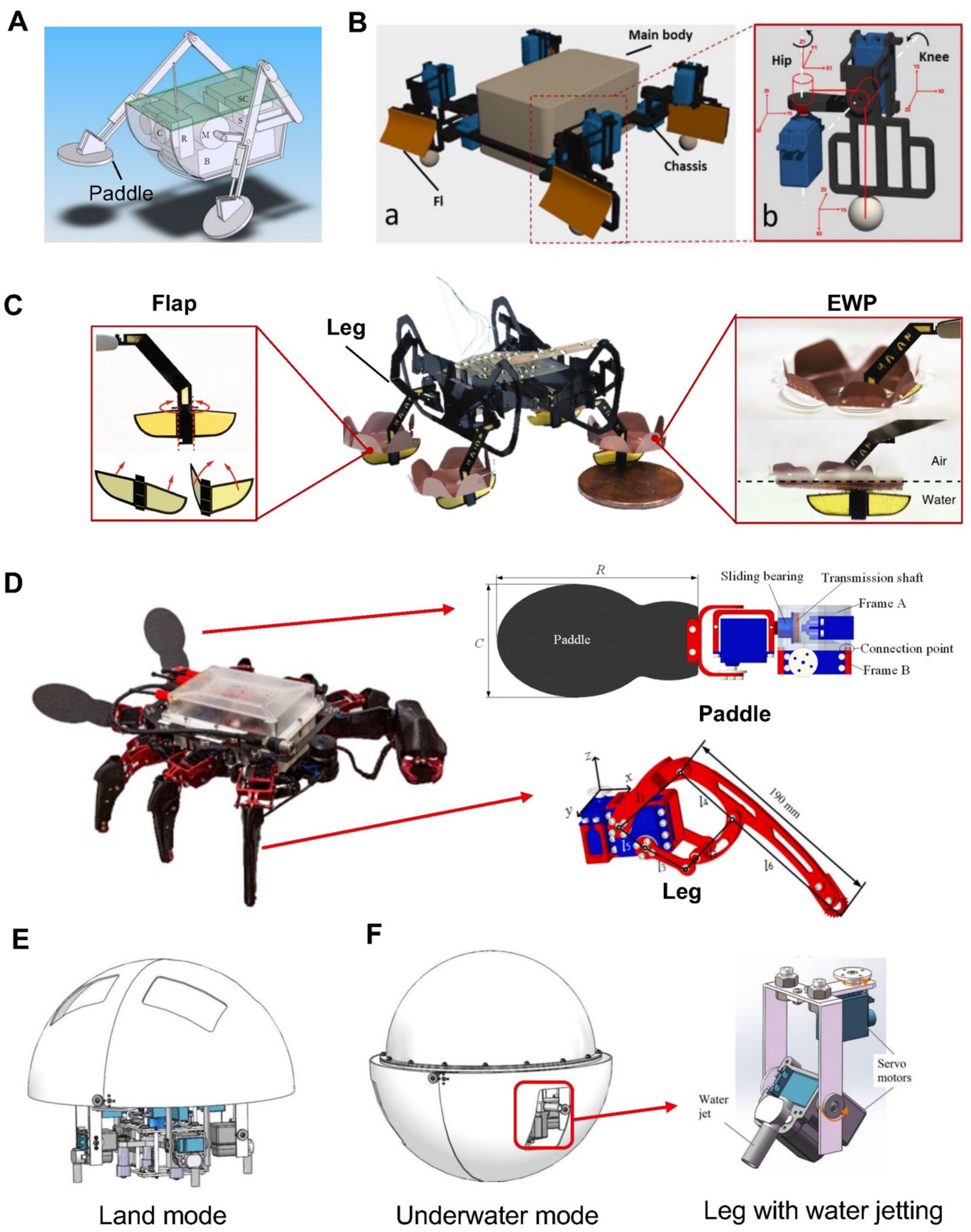

**Fig. 5. Legged-paddle and legged-water-jetting locomotion in amphibious robots. (A)** Schematic of a water-running robot inspired by basilisk lizards, using paddles as feet [4]. (**B)**

Quadruped amphibious robot with paddles integrated into its feet for aquatic propulsion [44]. (**C**) The design of a hybrid terrestrial-aquatic microrobot consists of an electrowetting pad (EWP) and flap on each leg [45]. **(D)** Hexapod amphibious bionic robot with two paddle-like structures acting as aquatic tails [46]. (**E**) Schematic of an amphibious spherical robot with retractable legs for land locomotion. **(F)** Water-jetting thrusters are installed at the ends of the robot's four legs for aquatic propulsion [47].

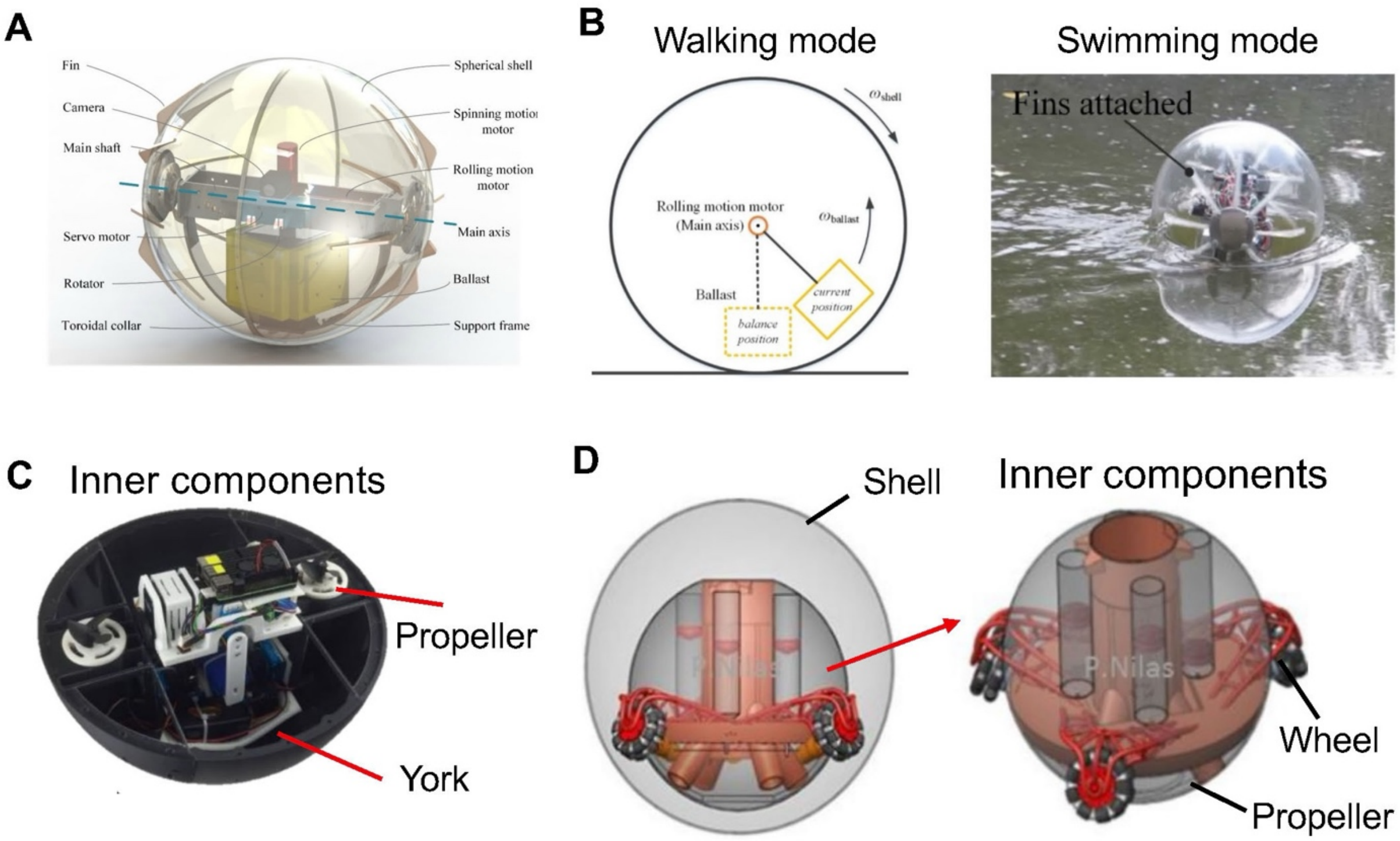

**Fig. 6. Spherical-rolling locomotion in amphibious robots. (A)** Schematic of an amphibious spherical robot equipped with assistant fins for enhanced propulsion [48]. (**B)** The transition between walking and swimming modes of the spherical robot. (**C–D)** Schematic illustrations of amphibious robots utilizing spherical-rolling and propeller locomotion for efficient land and water movement [49, 50].

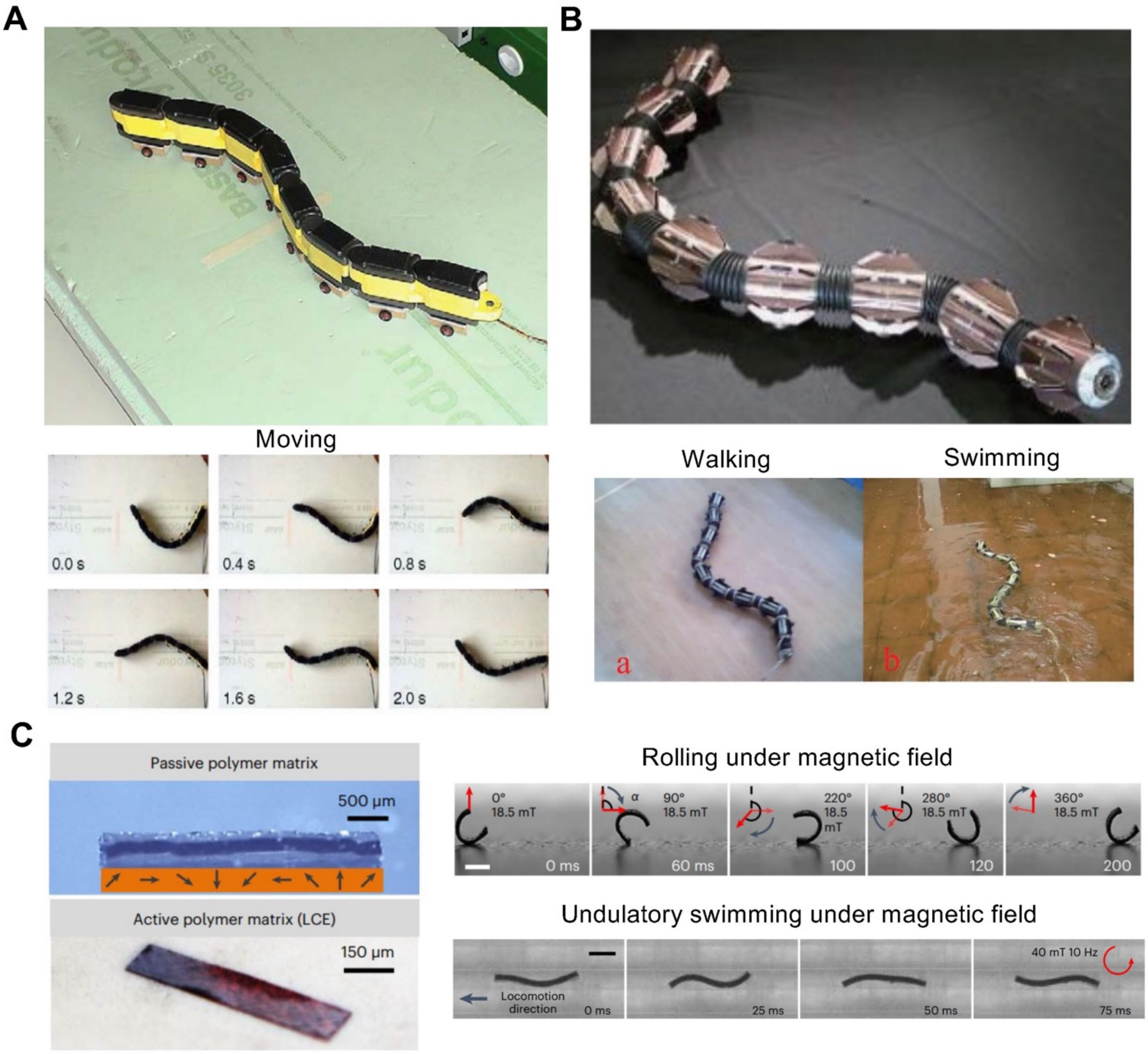

**Fig. 7. Body-undulation locomotion in amphibious robots. (A)** Schematic of the snake-like robot AmphiBot I utilizing body undulation for locomotion [51]. (**B.** Schematic of the snake-)like robot ACM-R, capable of walking and swimming through undulatory motion [1]. (**C**) Small-scale magnetic soft-body robot demonstrating rolling and undulatory swimming under an external magnetic field [53].

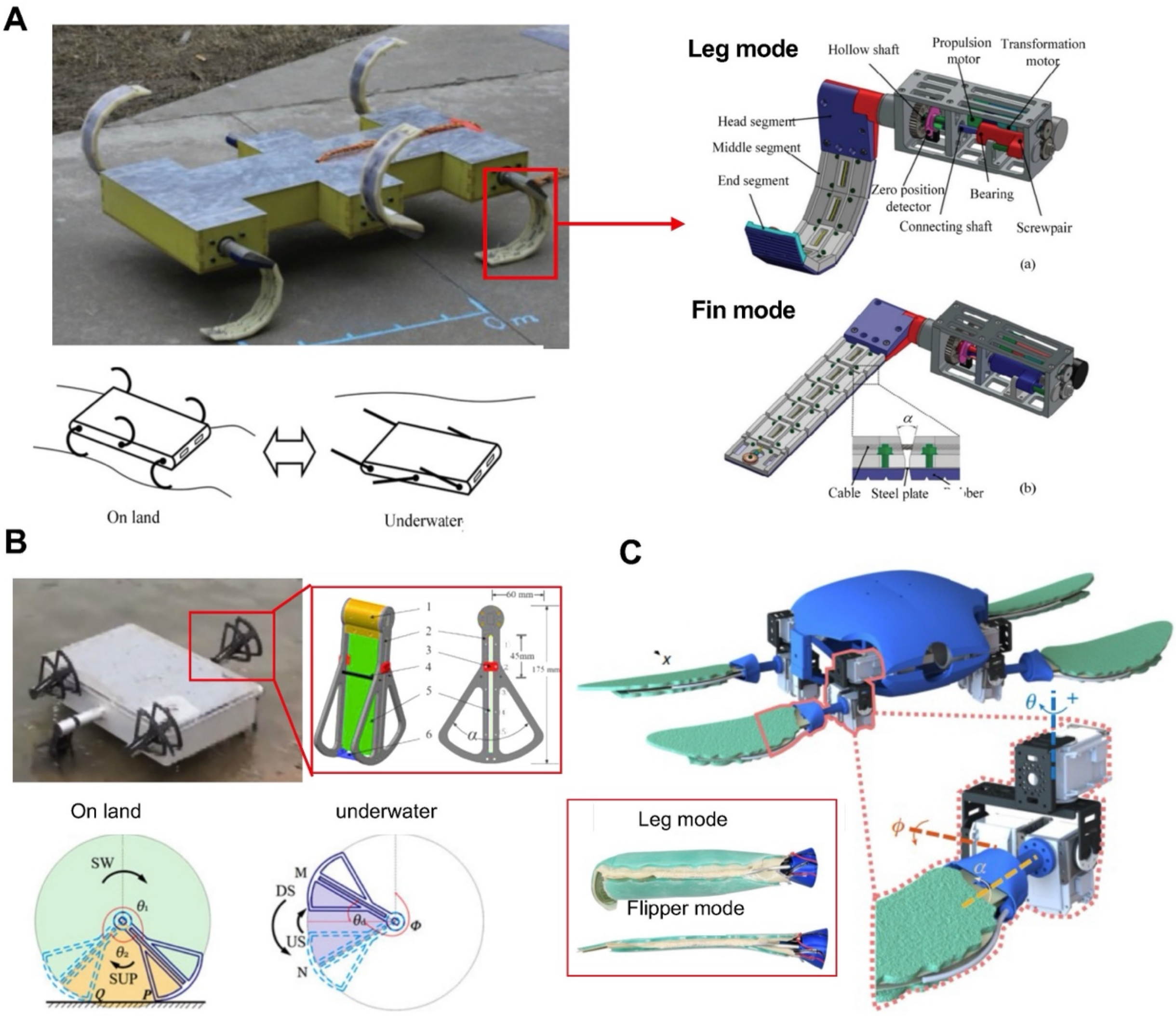

**Fig. 8. Flipper-leg locomotion in amphibious robots. (A)** Schematic of AmphiHex-I, where the robot's leg transitions between leg and fin modes for walking and swimming [54]. (**B)** Schematic of AmphiHex-II, integrating paddles and legs, with mode switching controlled by CPG signals [6]. (**C)** Turtle-inspired amphibious robot utilizing flipper-leg locomotion, transitioning between modes via inflation and deflation [3].

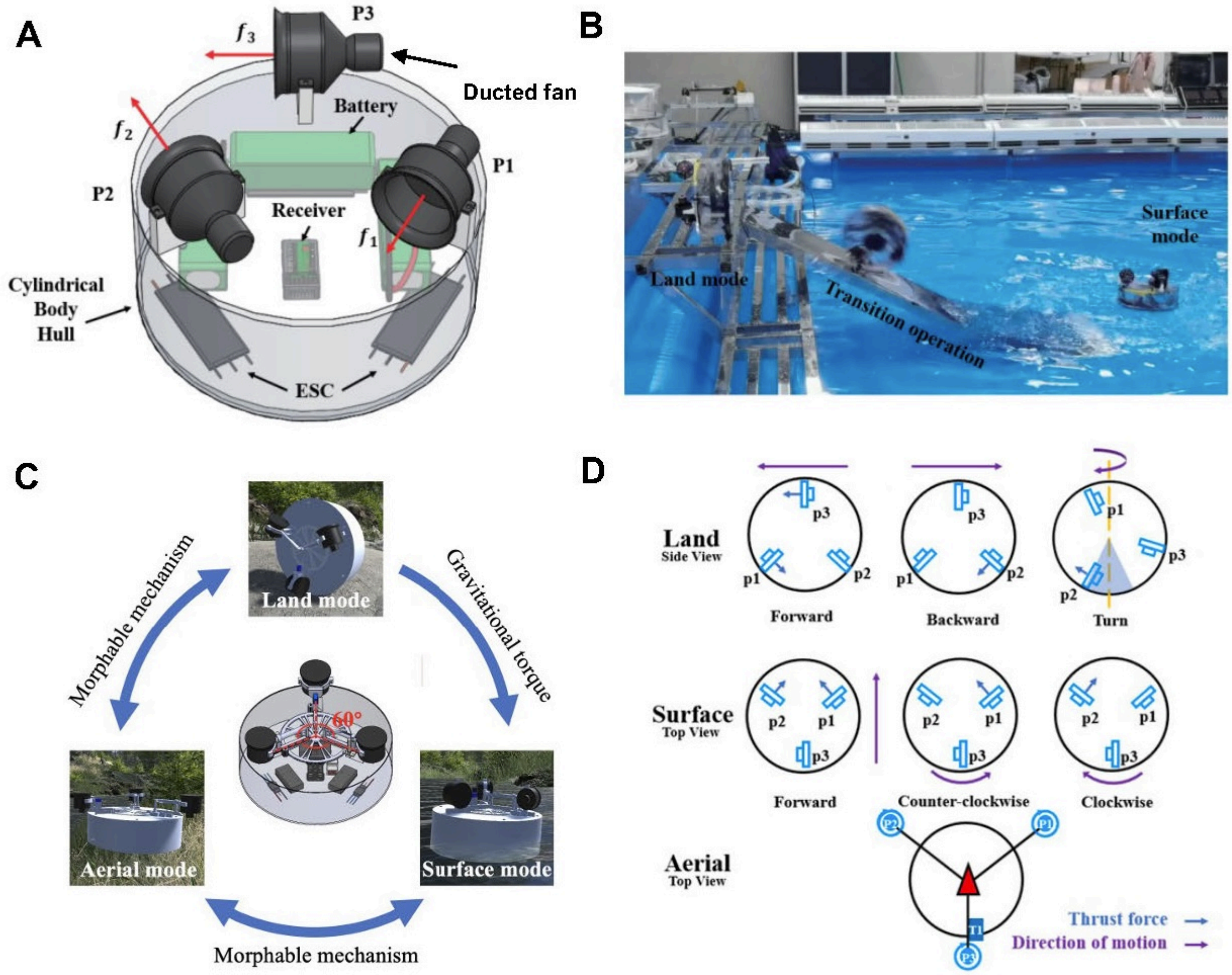

**Fig. 9. Rolling-hovercraft locomotion in amphibious robots. (A)** Schematic of a rolling-hovercraft amphibious robot with land and surface modes [7]. **(B)** In land mode, the robot advances by rolling and flips to switch modes by adjusting its center of gravity. In surface mode, it operates as a hovercraft. **(C)** Schematic of a triphibian robot capable of operating on land, water, and in the air [55]. **(D)** Illustration of the three motion modes—rolling, surface hovering, and aerial flight—of the triphibian robot.

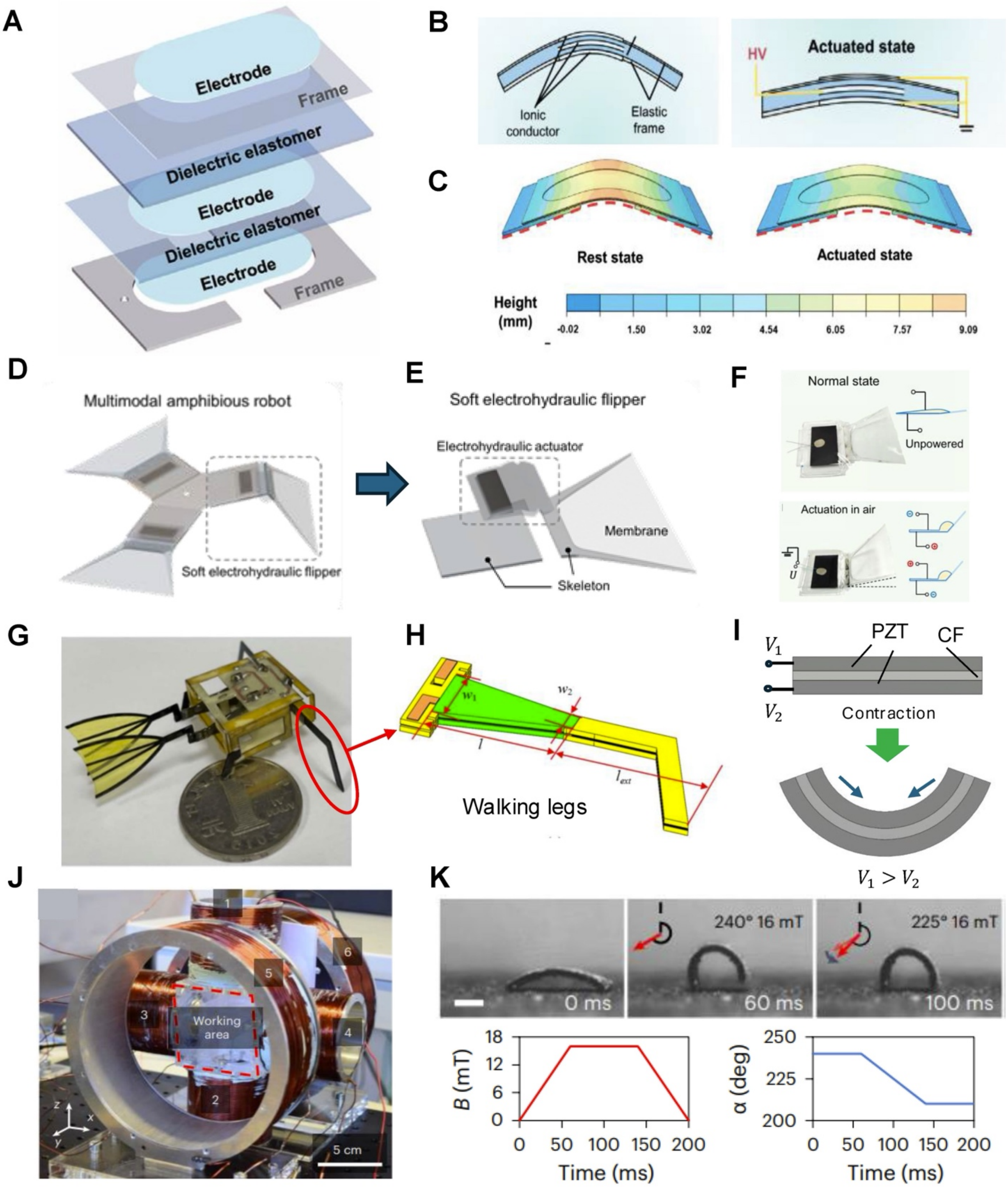

**Fig. 10. Actuation mechanisms in amphibious robots. (A)** Schematic of a dielectric elastomer actuator used in an amphibious robot [59]. (**B–C)** The application of voltages is used to bend the robot's body for locomotion. **(D–E)** Schematic of a Electrohydraulic Actuator integrated into an multimodal amphibious robot [61]. **(G–H)** Schematic of a piezoelectric polymer integrated into an amphibious robot [41]. **(I)** Differential voltage application to piezoelectric plates on each side of a bimorph mechanism causes asymmetric contraction, bending the robot's leg. **(J)** Electromagnetic coil system enabling magnetic actuation [53]. **(K)** Realization of walking modes under the influence of a magnetic field.

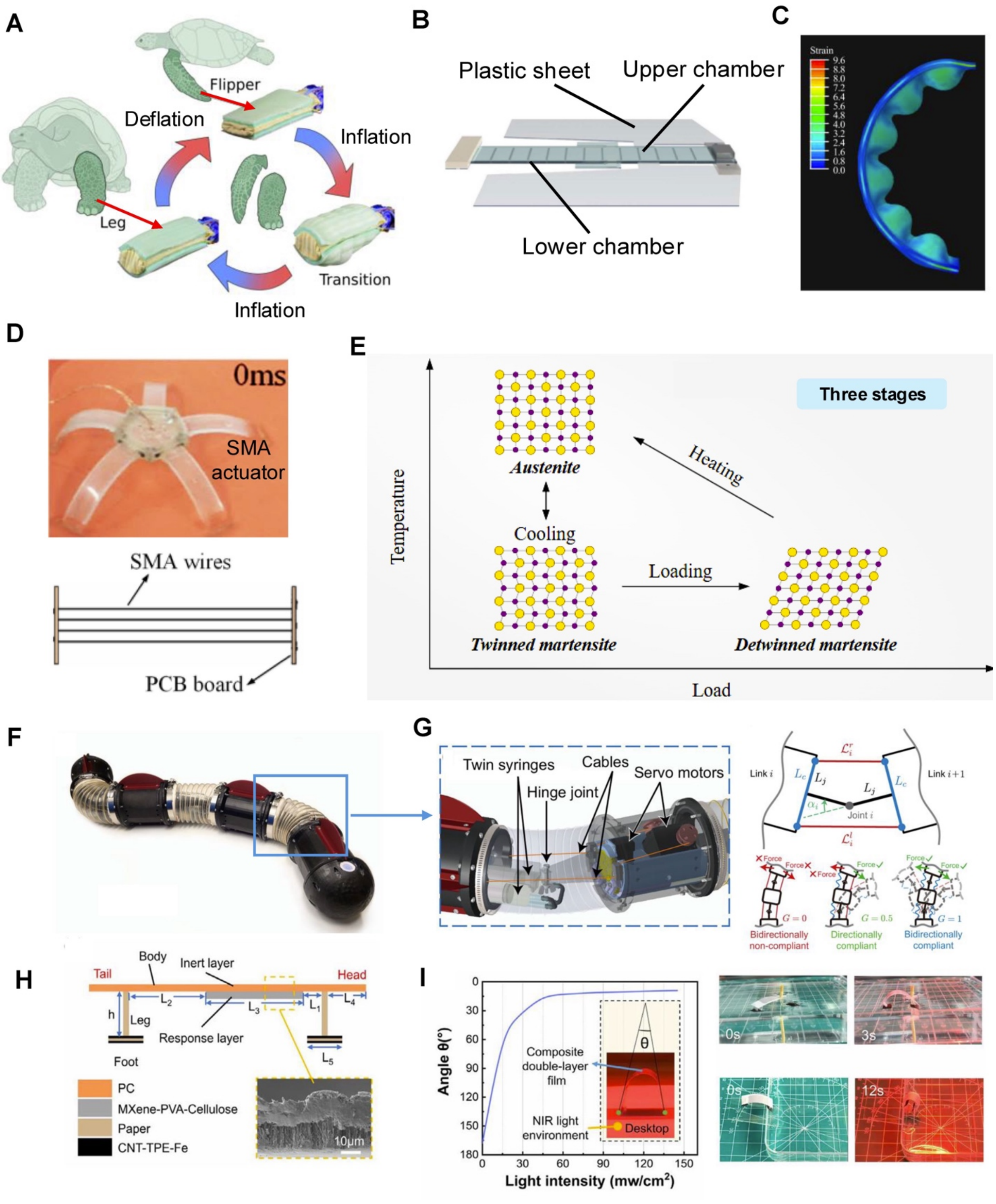

**Fig. 11. Soft actuation mechanisms in amphibious robots. (A)** Schematic of an amphibious robot driven by two Fluidic Elastomer Actuators (FEAs), enabling transitions between leg and flipper modes in a turtle-inspired robot [12]. (**B–C)** FEA is integrated into the body of an amphibious robot, enabling motion and flexibility for both terrestrial and aquatic environments [13]. (**D–E)** Schematic of a starfish-inspired amphibious robot with Shape Memory Alloy (SMA) actuators illustrating its working principle for adaptive locomotion [63]. **(F–G)** Schematic of a cable-driven limbless robot [64]. **(H–I)** A light-driven amphibious robot [65].

**Table 1.** Pros and Cons of Different Terrestrial Locomotion Approaches in Amphibious Robots.

| Locomotion Approach | Pros | Cons | Ref |
|---|---|---|---|
| Wheel/track Locomotion | • High energy efficiency<br>• Simple design<br>• High load-carrying capacity<br>• High speed on level terrains<br>• Easy to control and steer | • Limited adaptability to uneven or rough terrains<br>• Reduced efficiency on steep gradients<br>• Dependency on well-maintained surfaces | [14] |
| Legged Locomotion | • Superior adaptability to complex terrains<br>• Excellent obstacle-surmounting ability<br>• Versatile for different environments | • Lower speed compared to wheels/tracks<br>• Complex structure<br>• Prone to instability on loose surfaces<br>• Energy inefficient | [15] |
| Spherical rolling | • Omnidirectional movement<br>• Simple and compact design<br>• Reduced damage risks | • Difficulty in precise control<br>• Limited suitability for steep or uneven terrains<br>• Limited load-carry capacity<br>• Energy inefficient | [16] |

**Table 2.** Pros and Cons of Different Aquatic Locomotion Approaches in Amphibious Robots.

| **Locomotion Approach** | **Pros** | **Cons** | **Ref** |
|---|---|---|---|
| Paddle Locomotion | • Simple mechanical design<br>• Simple and efficient for propulsion<br>• Effective for low-speed, stable swimming | • Energy inefficient for sustained operation<br>• Reduced efficiency at high speeds<br>• Noisy operation | [23] |
| Fin Locomotion | • Energy-efficient for low-speed swimming<br>• Good maneuverability in confined spaces<br>• Quiet and less disruptive to the environment | • Limited speed<br>• Low load-carrying capacity<br>• Complexity in control mechanisms<br>• Inefficient in environments with strong currents or for high speeds<br>• | [24] |
| Water jetting | • High burst speeds<br>• Excellent maneuverability<br>• Effective in open water and high-speed environments | • High COT<br>• Complex mechanical system<br>• Requires precise design to minimize turbulence | [25] |
| Propeller Locomotion | • Simple design<br>• High speed and energy efficiency<br>• Well-suited for sustained underwater movement | • Poor maneuverability in tight or complex spaces<br>• Noise generation<br>• Creates turbulence in water | [26] |

**Table 3**. Pros and Cons of Different Actuation Methods in Amphibious Robots.

| Actuator | Pros | Cons | Typical Amphibious robot |
|---|---|---|---|
| Motors (Rigid Actuators) | • High precision<br>• Strong load-carrying capacity<br>• Reliable for robust tasks | • Heavy and bulky<br>• Limited flexibility & adaptability<br>• High power consumption | AQUA Robot [57]. ARGO Amphibious UGV [58]. |
| Dielectric Elastomer Actuators (DEA) | • Lightweight and flexible<br>• Capable of large deformations<br>• Energy-efficient for soft movements | • Low force output<br>• Need high voltages<br>• Limited durability in harsh environments | DEA-based Soft Robot [59]. |
| Electrohydraulic Actuator (EHA) | • Soft, compliant, silent actuation with large strain.<br>• Fast responsive | • High voltage<br>• Dielectric breakdown<br>• Safety constraint | Multimodal amphibious robot [61]. |
| Piezoelectric Polymer Actuators | • High precision and fast response<br>• Compact and lightweight<br>• Effective for small-scale robots | • Limited range of motion<br>• Requires amplification for larger movements<br>• Expensive materials | SAM Robot [41]. |
| Magnetic Actuators | • Compact and fast response<br>• Suitable for confined spaces<br>• Available for non-contact applications | • Requires external magnetic fields<br>• Limited range of applicability<br>• Sensitivity to magnetic interference | Magnetically controlled soft robot [53]. |
| Fluidic Elastomer Actuators (FEA) | • High flexibility<br>• High adaptability<br>• Smooth transitions between modes | • Low energy efficiency<br>• Require complex pumping systems | Soft-Robotics Amphibian Robosoft Amphibious Robot [13]. |
| Shape Memory Alloys (SMA) | • Compact and lightweight<br>• Large deformation<br>• Effective for underwater cooling scenarios | • Slow response time (cooling)<br>• Limited cyclic lifespan<br>• Low energy efficiency | Starfish robot [63]. |
| Cable-driven | • High force output<br>• Precise motion<br>• Robust to pressure | • Friction, backlash, and hysteresis<br>• Antagonistic pairs needed | AquaMILR+ [64]. |
| Light-driven | • Remotely control<br>• Environment-friendly stimuli<br>• Simple structure | • High CoT<br>• Requires high irradiance<br>• Challenged to scale | Light-driven amphibious robot[65]. |

**Table 4.** Control Strategies in Amphibious Robots.

| Control strategy | Pros | Cons | Applications |
|---|---|---|---|
| PID Control | • Simple and reliable<br>• Effective for maintaining stability<br>• Low computational demand | • Limited adaptability in dynamic or nonlinear environments<br>• Requires manual tuning | Used in robots like AmphiBot-I for stable undulatory motion during swimming and crawling [51]. |
| Model Predictive Control (MPC) | • Optimizes control inputs over a prediction horizon<br>• Suitable for complex dynamic systems | • High computational cost<br>• Challenging for real-time applications | Effective in optimizing fin and limb movements for energy-efficient locomotion[68-70]. |
| Finite State Machine (FSM) | • Structured and predictable<br>• Simplifies mode-switching tasks | • Lacks adaptability in unstructured environments<br>• Heavily reliant on predefined logic | Used in robots with well-defined environmental transitions, such as walking-to-swimming modes [71]. |
| Reinforcement Learning (RL) | • Adaptive to complex, uncertain environments<br>• Can optimize locomotion through learning | • Requires extensive training<br>• High computational and data demands | Applied in Amphibious Wheel-Legged Robot (AWLR) for optimizing mode transitions [72]. |
| Bio-Inspired Control | • Mimics natural systems for efficient adaptation<br>• Intuitive and flexible | • Limited performance in engineered environments<br>• Complexity in replicating behaviors | Applied in salamander-like robots for undulatory and legged locomotion [73]. |
| Behavior-Based Control | • Modular and easy to implement<br>• Effective for multitasking (e.g., path-following, obstacle avoidance) | • Behavior conflicts can lead to inefficient or unpredictable responses | Suitable for modular tasks in dynamic environments [47]. |
| Adaptive Control | • Adjusts parameters in real-time<br>• Robust in dynamic environments | • Complexity in algorithm design<br>• Requires extensive tuning and testing | Used for robots operating in unpredictable terrains or turbulent waters [74]. |

**Table 5.** Sensing and Control Strategies for Mode Switching in Amphibious Robots.

| **Sensing Technologies** | **Decision-making trigger** | **Control Strategy** | **Amphibious Robot/System** |
|---|---|---|---|
| Water sensor | Detects water presence above a threshold | Activate buoyancy devices or propellers; disable terrestrial actuators | AmphiBot [51].<br>AQUA Robot [57]. |
| Depth sensor | Detects depth increase | The transition from terrestrial legs to swimming fins or paddles | Salamandra Robotic [73]. |
| IMU | Detects slippage, tilt, or unstable land locomotion | Adjust posture; deploy underwater propulsion systems | Salamandra Robotic [73].<br>AQUA Robot [57]. |
| Visual sensor | Identifies terrain | Deploy appropriate locomotion mechanisms | Amphibious Hexapod Robot [46]. |
| Tactile sensor | Detects loss of traction on land | Switch to swimming or stabilize for land locomotion | RHex-based Amphibious Robot [54]. |